\documentclass[a4paper,twoside]{article}

\usepackage{epsfig}
\usepackage{subcaption}
\usepackage{calc}
\usepackage{amssymb}
\usepackage{amstext}
\usepackage{amsmath}
\usepackage{amsthm}
\usepackage{multicol}
\usepackage{pslatex}
\usepackage{apalike}
\usepackage{algorithm2e}
\usepackage[bottom]{footmisc}
\usepackage{SCITEPRESS}     
\usepackage{tikz}
\usetikzlibrary{
	shapes.geometric,
	arrows.meta,
	positioning,
	fit,
	backgrounds
}
\usepackage{booktabs}

\begin{document}

\title{Chaos-SSL: An Attention-Based Self-Supervised Learning Framework with Chaotic Transformation for Medical Image Classification}

\author{\authorname{Joao Batista Florindo\sup{1}\orcidAuthor{0000-0002-0071-0227}}
	\affiliation{\sup{1}Institute of Mathematics, Statistics and Scientific Computing, University of Campinas, Rua Sergio Buarque de Holanda, 651, Campinas, Brazil}
	\email{florindo@unicamp.br}
}


\abstract{Self-Supervised Learning (SSL) has emerged as a powerful paradigm to mitigate the reliance on large, annotated datasets, a common bottleneck in medical image analysis. However, standard SSL methods, which rely on simple geometric and color augmentations, may fail to capture the fine-grained, complex textural details necessary for classifying subtle pathologies. This paper introduces Chaos-SSL, a novel two-stage framework for medical image classification. In the first stage, we propose a new self-supervised pre-training strategy that leverages 1D chaotic maps (Logistic, Tent, and Sine) as a complex, non-linear augmentation for contrastive learning. We hypothesize that these chaotic transformations create ``harder'' and more semantically-rich views, forcing a network to learn robust representations of fine-grained medical textures. In the second stage, we introduce an attention-based fusion model that dynamically combines the specialized features from our Chaos-SSL model with the general-purpose features of a larger, ImageNet-pre-trained model. We validate our method on two public datasets: ISIC 2018 (skin lesions) and APTOS 2019 (diabetic retinopathy). Our results demonstrate that the Chaos-SSL model pre-trained with a Tent map for 30 epochs, followed by attention fusion, achieves performance fully competitive with the state-of-the-art, yielding an accuracy of 0.9261 on ISIC 2018 and 0.8726 on APTOS 2019. This significantly outperforms existing SSL methods, including several recent approaches.}

\onecolumn \maketitle \normalsize \setcounter{footnote}{0} \vfill

\section{\uppercase{Introduction}}
Deep learning, particularly Convolutional Neural Networks (CNNs), has become the cornerstone of modern medical image analysis, achieving human-level performance on tasks such as classification, segmentation, and detection \cite{ref_1}. However, this success is heavily predicated on the availability of large-scale, high-quality, and expertly annotated datasets. In the medical domain, acquiring such annotations is exceptionally expensive, time-consuming, and requires specialized expertise, leading to a ``data scarcity'' bottleneck.

Self-Supervised Learning (SSL) has emerged as a compelling solution to this problem \cite{ref_1}. By designing pretext tasks that do not require human labels, SSL methods can learn rich, generalizable feature representations from vast quantities of unlabeled data. The dominant SSL paradigm, contrastive learning, (e.g., SimCLR \cite{ref_simclr} and MoCo \cite{ref_moco}), trains a model to pull representations of positive pairs (e.g., two augmented views of the same image) closer together in an embedding space, while pushing negative pairs (views from different images) apart.

The efficacy of contrastive learning is fundamentally dependent on the quality of its data augmentations. Standard augmentations—such as random crops, flips, color jitter, and grayscale—are designed to be ``domain-preserving'' and teach invariance to simple geometric or color changes. While effective for natural images, these augmentations may be insufficient for medical imaging, where pathological information is often encoded in subtle, fine-grained textural differences rather than in global object shape or color \cite{ref_3}.

This limitation has spurred research into more sophisticated SSL strategies. Some focus on pretext tasks, such as solving Jigsaw puzzles, as seen in the recent FG-SSL paper by Park \& Ryu (2024), to learn spatial relationships \cite{ref_jigsaw}. Others explore more complex, deformable augmentations \cite{ref_4}. However, a significant gap remains in developing augmentation strategies that can specifically model the complex, non-linear, and semi-chaotic nature of biological textures and disease patterns.

This paper poses a new research question: Can we create a more powerful SSL framework by replacing simple, stochastic augmentations with complex, deterministic, non-linear transformations derived from chaos theory?

We propose \textbf{Chaos-SSL}, a novel framework that uses 1D chaotic maps as a data augmentation strategy for contrastive learning. Our hypothesis is twofold:
\begin{enumerate}
	\item By applying chaotic maps (Logistic, Tent, and Sine) pixel-wise to an image, we create a ``hard'' positive view that exhibits complex, non-linear distortions. To minimize the contrastive loss, the model is forced to learn features that are invariant to this chaotic distortion, thereby capturing essential, fine-grained textural information.
	\item A single model cannot optimally capture all necessary features. Therefore, we hypothesize that a final model, which uses an ``attention mechanism'' to fuse the specialized, texture-aware features of our Chaos-SSL model with the robust, context-aware features of a large ImageNet-trained model, will achieve state-of-the-art performance.
\end{enumerate}

To validate this, we implement a two-stage pipeline. First, we pre-train a ConvNeXt-Tiny model using our chaotic contrastive method. Second, we fuse its features with a ConvNeXt-Large model using an attention-based ensemble. We test this on the ISIC 2018 and APTOS 2019 datasets and compare our results against the state-of-the-art, including the recent FG-SSL \cite{ref_jigsaw} method. Our contributions are:
\begin{itemize}
	\item The first (to our knowledge) application of chaotic maps as an explicit data augmentation technique for self-supervised contrastive learning.
	\item The design of a ChaoticTransform class compatible with standard deep learning pipelines.
	\item An attention-based ensemble model that effectively fuses generalist (ImageNet) features with specialist (Chaos-SSL) features.
	\item A comprehensive empirical study showing our method significantly outperforms prior approaches on two challenging medical classification benchmarks.
\end{itemize}

This paper is structured as follows: Section 2 reviews related work. Section 3 details our proposed methodology. Section 4 presents our experimental results. Section 5 discusses the implications of our findings, and Section 6 concludes the paper.

\section{\uppercase{Related Work}}
Our work is situated at the intersection of three active research areas: self-supervised learning in medical imaging, data augmentation, and feature fusion.

\subsection{Self-Supervised Learning in Medical Imaging}
SSL has been a major focus of the medical imaging community since 2020 \cite{ref_1}. It is a promising solution for the challenges of small and imbalanced datasets, which are common in medicine \cite{ref_5}. The field has evolved from early pretext tasks (e.g., rotation prediction) to more sophisticated contrastive and non-contrastive methods.

Recent works have focused on adapting SSL to the specific properties of medical data. For instance, FG-SSL \cite{ref_jigsaw} adapts the Jigsaw puzzle pretext task to medical images. It uses a ``progressive Jigsaw puzzle'' strategy, arguing that learning to piece together patches of different granularities helps the model learn fine-grained features. Their method, combined with BarlowTwins (a non-negative contrastive loss) \cite{ref_jigsaw_barlow}, demonstrates strong performance. Other recent approaches include leveraging self-distillation (e.g., DINOv2) \cite{ref_6} and combining contrastive learning with masked autoencoders \cite{ref_7}. These works confirm that the type of pretext task or learning signal is a critical design choice.

\subsection{Chaos-Based Data Augmentation}
While the SSL framework is important, the data augmentation pipeline is arguably one of its most critical components. Recent research, such as a 2024 study on segmentation \cite{ref_4}, has shown that specialized, deformable data augmentations can outperform standard contrastive learning. This suggests that creating more ``difficult'' or ``semantically relevant'' transformations is key.

Our work draws inspiration from the intersection of chaos theory and deep learning. Chaos theory provides a mathematical foundation for modeling complex, non-linear, and deterministic systems. Recent studies, such as ``Neurochaos Learning'' (2025) \cite{ref_8}, have begun to use chaotic maps (e.g., Logistic, Sine, Tent) directly within the learning paradigm, harnessing their complex dynamics. Others have explored using chaotic maps for federated learning encryption \cite{ref_9}. However, the application of chaotic maps as a \textit{data augmentation} technique for contrastive visual learning remains a novel and unexplored area. Our work fills this gap.

\subsection{Attention-Based Feature Fusion}
It is well-established that features from models pre-trained on different tasks (e.g., general-purpose ImageNet vs. domain-specific SSL) are often complementary. The challenge lies in how to combine them. Simple concatenation or averaging of logits is often suboptimal.

Recent literature (2023-2025) strongly advocates for more sophisticated fusion techniques. A 2023 review of medical image fusion highlights the trend toward deep learning-based methods \cite{ref_10}. A 2025 study proposed a ``Multi-Stage Feature Fusion Network'' (MSFF) that uses ``Context Modulated Attention'' to fuse local and global features for medical image classification \cite{ref_11}. Another 2025 paper introduced an ``Ensemble Fusion AI'' (EFAI) that intelligently concatenates features from multiple models \cite{ref_12}. This trend confirms our hypothesis that a learnable, attention-based fusion mechanism is a state-of-the-art approach for combining feature streams from our two distinct backbone models.

\section{\uppercase{Methodology}}
Our proposed framework, Chaos-SSL, consists of a multi-stage pipeline designed to generate and fuse specialized feature representations. This section details the components.

\subsection{Stage 1: Chaotic-Contrastive Pre-training (Chaos-SSL)}
The first stage pre-trains a ConvNeXt-Tiny encoder using a novel self-supervised contrastive task. The core of this task is a new augmentation, the ChaoticTransform.

\subsubsection{The ChaoticTransform}
We designed a custom torchvision transform, $T_{chaos}$, that applies a 1D chaotic map $f(x)$ pixel-wise for a number of iterations $k$, where $k \sim \mathcal{U}[k_{min}, k_{max}]$. The input image tensor $I$, assumed to be normalized $I \in [0, 1]^{C \times H \times W}$, is treated as the initial state $I^{(0)}$. To ensure numerical stability, we clamp the input: $I^{(0)} \leftarrow \text{clamp}(I, \epsilon, 1-\epsilon)$. The map is then iterated:
$$ I^{(n+1)} = f(I^{(n)}) \quad \text{for } n = 0, \dots, k-1 $$
The final transformed image is $I_{chaos} = I^{(k)}$.

We implement three distinct chaotic maps:
\begin{enumerate}
	\item \textbf{Logistic Map:} $f(x) = r \cdot x (1 - x)$, with $r = 3.99$.
	\item \textbf{Tent Map:} $f(x) = \mu \cdot \min(x, 1 - x)$, with $\mu = 2.0$.
	\item \textbf{Sine Map:} $f(x) = r \cdot \sin(\pi \cdot x)$, with $r = 1.0$.
\end{enumerate}
In our experiments, $k$ is randomly sampled from $\mathcal{U}[1, 5]$.

\subsubsection{SSL Framework}
We adopt a SimCLR-style contrastive framework. For each image $I$ in a batch, we generate two views:
\begin{itemize}
	\item \textbf{View 1 ($v_1$):} $v_1 = T_{std}(I)$, where $T_{std}$ is a composition of standard augmentations (RandomHorizontalFlip, ColorJitter, RandomGrayscale, GaussianBlur).
	\item \textbf{View 2 ($v_{chaos}$):} $v_{chaos} = T_{chaos}(T_{std}(I))$. The chaotic map is applied after the standard augmentations.
\end{itemize}

These views are fed into our ContrastiveModel, $M_{ssl}$, which consists of an encoder $E(\cdot)$ (a ConvNeXt-Tiny backbone) and a projector $P(\cdot)$ (a 3-layer MLP):
$$ z_1 = P(E(v_1)) \quad \text{and} \quad z_{chaos} = P(E(v_{chaos})) $$
The projections $z \in \mathbb{R}^{128}$ are used to compute the NT-Xent loss. For a positive pair $(z_i, z_j)$ from a mini-batch of size $N$ (resulting in $2N$ total vectors), the loss for vector $z_i$ is:
$$ \mathcal{L}_i = -\log \frac{\exp(\text{sim}(z_i, z_j) / \tau)}{\sum_{k=1, k \neq i}^{2N} \exp(\text{sim}(z_i, z_k) / \tau)} $$
where $\text{sim}(u, v) = \frac{u^\top v}{\|u\| \|v\|}$ is the cosine similarity and $\tau$ is the temperature parameter. The total loss for the batch is $\mathcal{L}_{NTX} = \frac{1}{2N} \sum_{i=1}^{2N} \mathcal{L}_i$.

\subsubsection{Training Details}
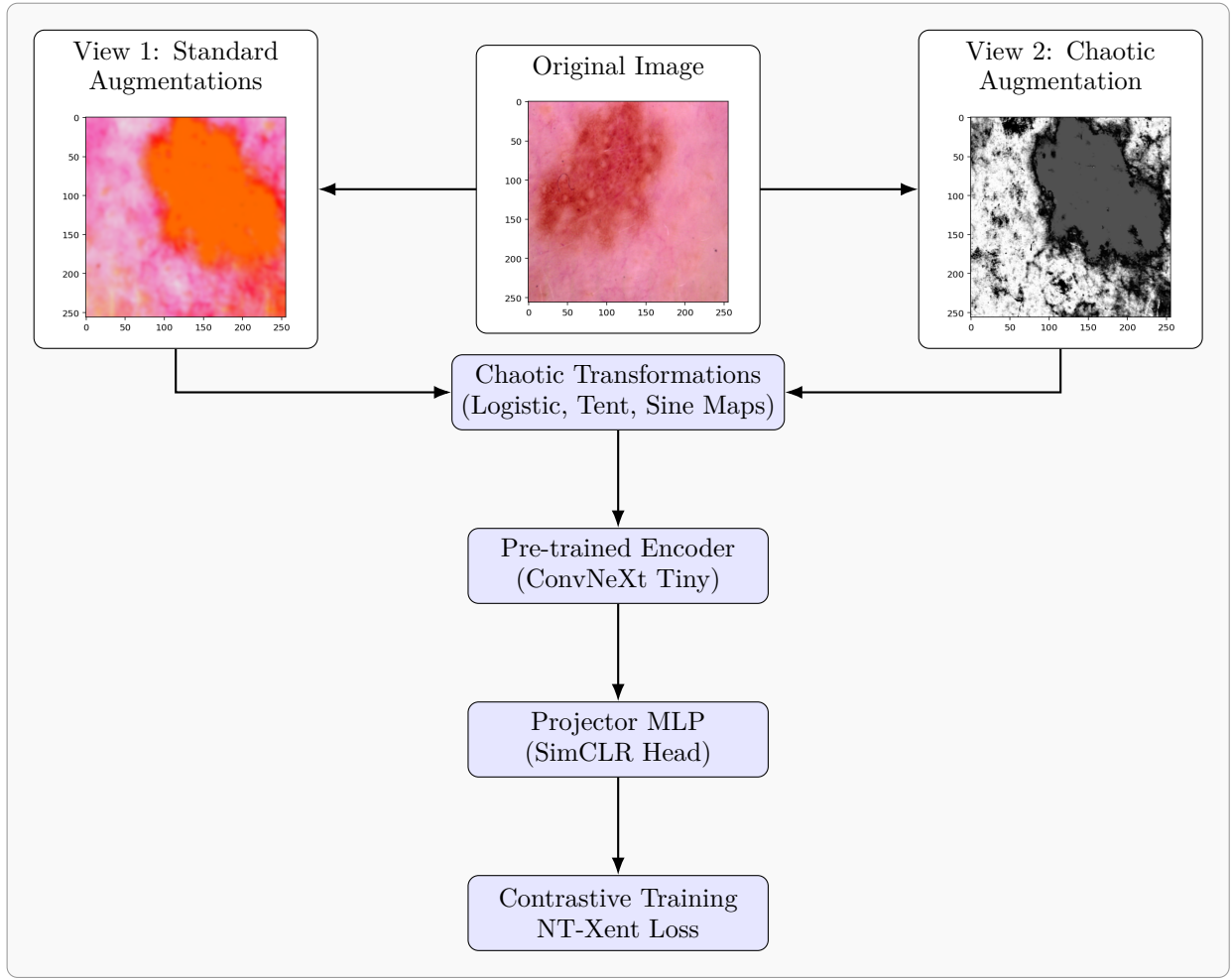
\begin{figure*}
	\centering
	\begin{tikzpicture}[
		node distance=1.3cm and 2cm,
		process/.style={rectangle, rounded corners, draw=black, fill=blue!10, align=center, minimum width=4cm, minimum height=1cm},
		data/.style={ellipse, draw=black, fill=orange!20, align=center, minimum width=3.5cm, minimum height=1cm},
		imagebox/.style={rectangle, draw=black, rounded corners, fill=white, inner sep=4pt, align=center, text width=3.5cm},
		line/.style={-Latex, thick},
		group/.style={draw=gray!80, rounded corners, inner sep=10pt, fill=gray!5}
		]
		
		
		\path (0,0) coordinate (base_coord);
		
		\node[imagebox, left=4cm of base_coord] (view1) {
			\textbf{View 1: Standard Augmentations}\\[4pt]
			\IfFileExists{aug_image.png}
			{\includegraphics[width=3.2cm]{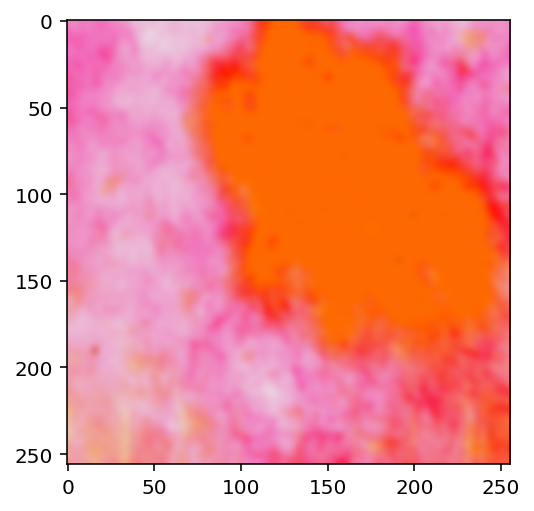}}
			{\fbox{\rule{0pt}{2cm}\hspace{3.2cm}}}
		};
		\node[imagebox, right=4cm of base_coord] (view2) {
			\textbf{View 2: Chaotic Augmentation}\\[4pt]
			\IfFileExists{chaotic_image.png}
			{\includegraphics[width=3.2cm]{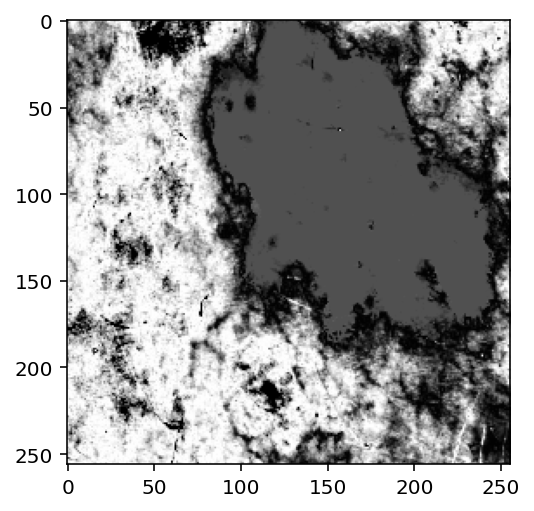}}
			{\fbox{\rule{0pt}{2cm}\hspace{3.2cm}}}
		};
		
		\node[process, below=2.2cm of base_coord] (chaotic) {Chaotic Transformations \\ (Logistic, Tent, Sine Maps)};
		\node[process, below=of chaotic] (encoder) {Pre-trained Encoder \\ (ConvNeXt Tiny)};
		\node[process, below=of encoder] (projector) {Projector MLP \\ (SimCLR Head)};
		\node[process, below=of projector] (contrastive) {Contrastive Training \\ NT-Xent Loss};
		
		\draw[line] (base_coord) |- (view1);
		\draw[line] (base_coord) |- (view2);
		
		\node[imagebox, at=(base_coord)] (base) {
			\textbf{Original Image}\\[4pt]
			\IfFileExists{base_image.png}
			{\includegraphics[width=3.2cm]{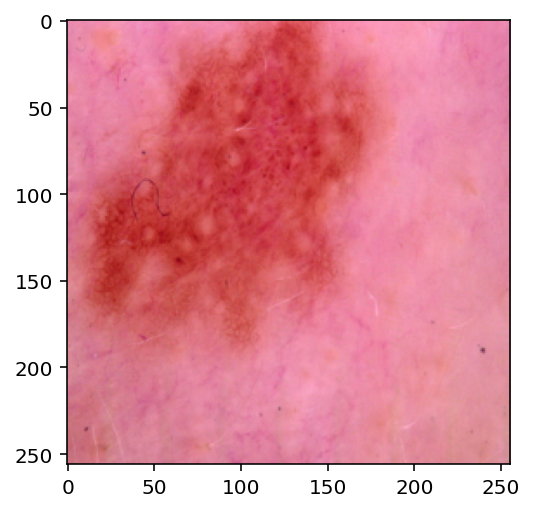}}
			{\fbox{\rule{0pt}{2cm}\hspace{3.2cm}}}
		};
		
		\draw[line] (view1.south) |- (chaotic.west);
		\draw[line] (view2.south) |- (chaotic.east);
		\draw[line] (chaotic) -- (encoder);
		\draw[line] (encoder) -- (projector);
		\draw[line] (projector) -- (contrastive);
		
		\begin{scope}[on background layer]
			<+
			\node[group,
			fit=(base)(chaotic)(encoder)(projector)(contrastive)(view1)(view2)]
			(sslgroup) {};
		\end{scope}			
	\end{tikzpicture}
	\caption{SSL Pre-training stage. An original image is split into two views: View 1 (Standard Augmentations) and View 2 (Chaotic Augmentation). Both views are fed into an encoder and a projector MLP , then trained using NT-Xent contrastive loss.}
	\label{fig:ssl}
\end{figure*}

We use the AdamW optimizer, with different learning rates $LR_E$ for the encoder and $LR_P$ for the projector.
Let $\Theta_E$ be the parameters of the encoder $E$ and $\Theta_P$ be the parameters of the projector $P$. The optimizer manages two parameter groups:
\begin{itemize}
	\item Group 1: $\Theta_E$, with $LR_E = 10^{-7}$
	\item Group 2: $\Theta_P$, with $LR_P = 10^{-3}$
\end{itemize}
The models are trained for 15 or 30 epochs. After training, $P$ is discarded and the weights $\Theta_E$ are saved. Figure \ref{fig:ssl} illustrates the overall pre-training process.

\subsection{Stage 2: Supervised Fine-Tuning of Base Models}
Before fusion, we fine-tune two base models on the target task using the standard cross-entropy loss:
$$ \mathcal{L}_{CE} = -\sum_{c=1}^{M} y_c \log(p_c) $$
where $M$ is the number of classes, $y_c$ is the one-hot encoded ground truth, and $p_c$ is the predicted probability from the softmax function: $p_c = \frac{\exp(l_c)}{\sum_{j=1}^{M} \exp(l_j)}$ for logits $l$.

\begin{itemize}
	\item \textbf{Model 1 (ImageNet-Large):} A ConvNeXt-Large model, pre-trained on ImageNet, is fine-tuned for 20 epochs. It uses a CosineAnnealingLR scheduler, where the learning rate $\eta_t$ at epoch $t$ is:
	$$ \eta_t = \eta_{min} + \frac{1}{2}(\eta_{max} - \eta_{min})\left(1 + \cos\left(\frac{T_{cur}}{T_{max}}\pi\right)\right) $$
	where $T_{cur}$ is the current epoch and $T_{max}$ is the total number of epochs.
	\item \textbf{Model 2 (Chaos-SSL-Tiny):} The ConvNeXt-Tiny encoder $E$ with weights $\Theta_E$ from Stage 1 is loaded, a new linear classifier head $C(\cdot)$ is attached ($M_{tiny} = C(E(\cdot))$), and the entire model is fine-tuned for 10 epochs.
\end{itemize}

\subsection{Stage 3: Attention-Based Feature Fusion}
The final stage combines the backbones of the two fine-tuned models, $B_1$ (Large) and $B_2$ (Tiny), using an attention-based feature fusion model.

\subsubsection{Fusion Architecture}
For an input $x$, we extract and concatenate features:
$$ f_1 = B_1(x) \in \mathbb{R}^{d_1} \quad f_2 = B_2(x) \in \mathbb{R}^{d_2} $$
$$ f_{concat} = [f_1, f_2] \in \mathbb{R}^{d_1 + d_2} $$
A Squeeze-and-Excite attention mechanism computes an importance vector $w_{attn}$:
$$ w_{attn} = \sigma\left( W_2 \cdot \delta\left( W_1 \cdot f_{concat} \right) \right) $$
where $W_1 \in \mathbb{R}^{d_{r} \times (d_1 + d_2)}$ and $W_2 \in \mathbb{R}^{(d_1 + d_2) \times d_{r}}$ are weights of linear layers, $d_r$ is the reduced dimension, $\delta$ is the ReLU function, and $\sigma$ is the Sigmoid function.
The features are re-weighted using the Hadamard product $\odot$:
$$ f_{attended} = f_{concat} \odot w_{attn} $$
A final classifier $C_{final}$ produces the logits:
$$ l_{final} = C_{final}(f_{attended}) = W_{class} \cdot f_{attended} + b_{class} $$

\subsubsection{Fusion Training}
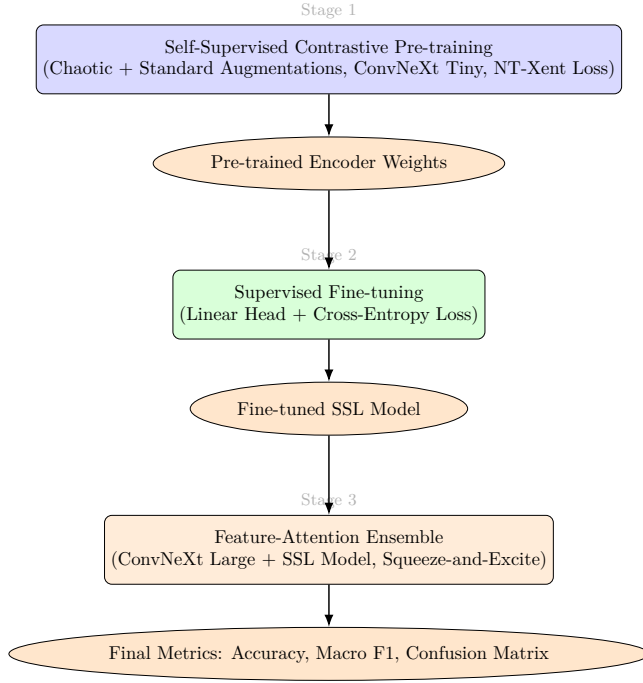
\begin{figure}
	\centering
	\scalebox{.7}{
		\begin{tikzpicture}[
			node distance=2cm and 2cm,
			stage/.style={rectangle, rounded corners, draw=black, fill=blue!10, text centered, minimum width=5cm, minimum height=1.3cm, align=center, font=\bfseries},
			process/.style={rectangle, rounded corners, draw=black, fill=green!10, align=center, minimum width=4cm, minimum height=1cm},
			result/.style={ellipse, draw=black, fill=orange!20, align=center, minimum width=4cm, minimum height=1cm},
			line/.style={-Latex, thick},
			every label/.append style={font=\footnotesize\bfseries, text=gray!60}
			]
			
			\node[font=\large\bfseries, align=center, text width=12cm, yshift=1cm]
			(title) {};
			
			\node[stage, below=1cm of title, fill=blue!15, label=above:{Stage 1}] (ssl)
			{Self-Supervised Contrastive Pre-training\\
				(Chaotic + Standard Augmentations, ConvNeXt Tiny, NT-Xent Loss)};
			
			\node[result, below=0.8cm of ssl] (sslout)
			{Pre-trained Encoder Weights};
			
			\node[stage, below=1.5cm of sslout, fill=green!15, label=above:{Stage 2}] (ft)
			{Supervised Fine-tuning\\
				(Linear Head + Cross-Entropy Loss)};
			
			\node[result, below=0.8cm of ft] (ftout)
			{Fine-tuned SSL Model};
			
			\node[stage, below=1.5cm of ftout, fill=orange!15, label=above:{Stage 3}] (ens)
			{Feature-Attention Ensemble\\
				(ConvNeXt Large + SSL Model, Squeeze-and-Excite)};
			
			\node[result, below=0.8cm of ens] (metrics)
			{Final Metrics: Accuracy, Macro F1, Confusion Matrix};
			
			\draw[line] (ssl) -- (sslout);
			\draw[line] (sslout) -- (ft);
			\draw[line] (ft) -- (ftout);
			\draw[line] (ftout) -- (ens);
			\draw[line] (ens) -- (metrics);
	\end{tikzpicture}}
	\caption{Overall architecture of the proposed three-stage methodology, starting with self-supervised contrastive pre-training, followed by supervised fine-tuning, and ending with a feature-attention fusion model that produces the final outcomes.}
	\label{fig:overall}
\end{figure}

This new model is trained with differential learning rates using AdamW. Let $\Theta_{B1}, \Theta_{B2}$ be the backbone parameters and $\Theta_{Head} = \{W_1, W_2, W_{class}, b_{class}\}$ be the new parameters. The optimizer manages:
\begin{itemize}
	\item Group 1: $\{\Theta_{B1}, \Theta_{B2}\}$, with $LR_{Backbone} = 10^{-6}$
	\item Group 2: $\Theta_{Head}$, with $LR_{Head} = 10^{-4}$
\end{itemize}
The fusion model is trained for 10 epochs. Fiure \ref{fig:overall} shows the high-level diagram of the steps involved in the proposed methodology.

\subsection{Datasets and Implementation}
We evaluate our method on two publicly available medical imaging datasets.
\begin{itemize}
	\item \textbf{ISIC 2018 \cite{ref_isic2018}:} A dataset of 10,015 dermoscopic images of skin lesions, classified into 7 categories.
	\item \textbf{APTOS 2019 \cite{ref_aptos2019}:} A dataset of 3,662 fundus images for diabetic retinopathy classification, with 5 severity grades.
\end{itemize}
All models were trained on an NVIDIA GPU (RTX 5080) using PyTorch, timm (for models), and `lightly` (for NT-Xent loss) libraries.

\section{\uppercase{Experimental Results}}
We now present the quantitative results of our Chaos-SSL framework, based on the outputs from our experimental framework. We analyze the impact of different chaotic maps and SSL training durations, and then compare our model against the state-of-the-art.

\subsection{Performance on APTOS 2019}
Table \ref{tab:aptos_results} shows the fusion model performance on the APTOS 2019 dataset. We observe some trends:
\begin{itemize}
	\item \textbf{Effect of Epochs:} Increasing the SSL pre-training from 15 to 30 epochs improves performance in most cases. For the Tent map, this yields a significant jump in F1-score from 0.7323 to 0.7601.
	\item \textbf{Effect of Map Type:} The choice of chaotic map also has some impact. The Tent map outperforms the Logistic and Sine maps at 30 epochs, achieving the highest accuracy (0.8726) and F1-score (0.7601). At 15 epochs, Sine map is slightly superior.
\end{itemize}

\begin{table}[!htpb]
	\centering
	\caption{Ensemble Performance on APTOS 2019 Dataset}
	\label{tab:aptos_results}
	\begin{tabular}{@{}lccc@{}}
		\toprule
		SSL & Chaotic & Accuracy & F1-Score \\
		 Epochs & Map & &  (Macro)\\
		\midrule
		15 & Logistic & 0.8644 & 0.7449 \\
		15 & Sine & 0.8681 & 0.7505 \\
		15 & Tent & 0.8599 & 0.7323 \\
		\midrule
		30 & Logistic & 0.8672 & 0.7484 \\
		30 & Sine & 0.8681 & 0.7483 \\
		\textbf{30} & \textbf{Tent} & \textbf{0.8726} & \textbf{0.7601} \\
		\bottomrule
	\end{tabular}
\end{table}

\begin{figure}[!htpb]
	\centering
	\includegraphics[width=0.9\columnwidth]{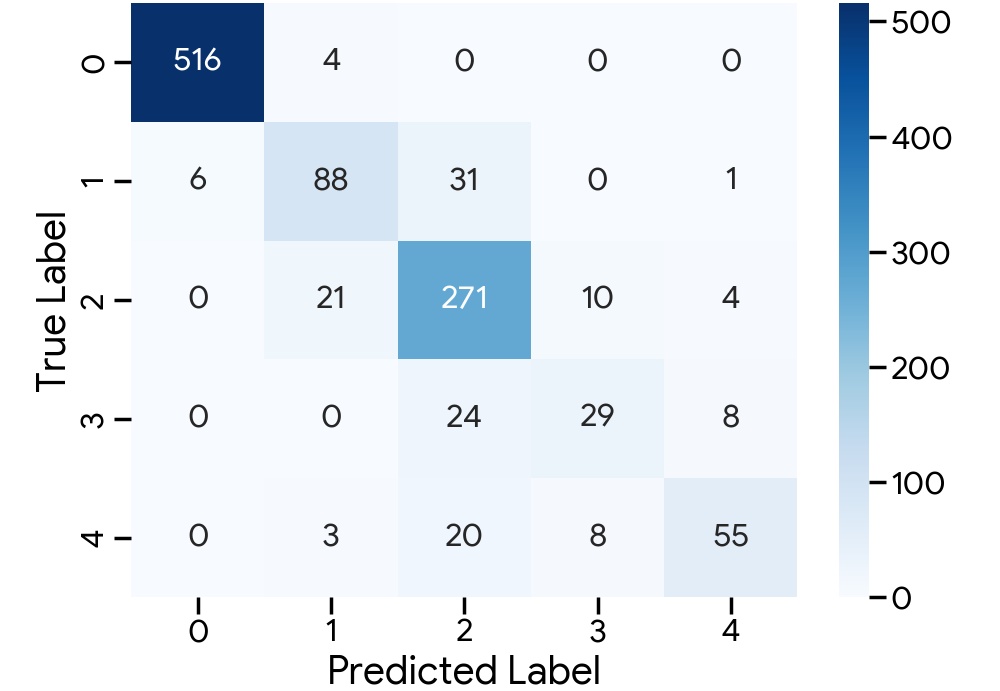}
	\caption{Confusion matrix for the best-performing model on the APTOS 2019 dataset (F1-Score: 0.7601). This corresponds to the 30-epoch Tent map SSL pre-training[cite: 394].}
	\label{fig:aptos_cm}
\end{figure}

\subsection{Performance on ISIC 2018}
The results on the more complex, 7-class ISIC 2018 dataset (Table \ref{tab:isic_results}) reinforce our findings.
\begin{itemize}
	\item \textbf{Effect of Epochs:} Again, 30 SSL epochs generally leads to better results than 15.
	\item \textbf{Effect of Map Type:} The \textbf{Tent map} once again proves to be the most effective, achieving the highest accuracy (0.9261) and F1-score (0.8706) when trained for 30 epochs. And, again, the Sine map is the best at 15 epochs.
\end{itemize}

\begin{table}[!htpb]
	\centering
	\caption{Ensemble Performance on ISIC 2018 Dataset}
	\label{tab:isic_results}
	\begin{tabular}{@{}lccc@{}}
		\toprule
		SSL & Chaotic & Accuracy & F1-Score \\
		Epochs & Map & & (Macro)\\
		\midrule
		15 & Logistic & 0.9198 & 0.8488 \\
		15 & Sine & 0.9245 & 0.8644 \\
		15 & Tent & 0.9195 & 0.8593 \\
		\midrule
		30 & Logistic & 0.9245 & 0.8641 \\
		30 & Sine & 0.9221 & 0.8545 \\
		\textbf{30} & \textbf{Tent} & \textbf{0.9261} & \textbf{0.8706} \\
		\bottomrule
	\end{tabular}
\end{table}

\subsection{Comparison with the Literature}
The primary goal of our method is to advance the state-of-the-art in medical SSL. We compare proposed model, in its configuration with 30 epochs and Tent map, against the results reported in \cite{ref_jigsaw}, which uses the same datasets.

As shown in Table \ref{tab:sota_comparison}, our Chaos-SSL framework outperforms all compared methods.
\begin{itemize}
	\item On \textbf{APTOS 2019}, our 0.8726 accuracy surpasses FG-SSL's 0.858 by 1.7\% and their F1-score is surpassed by 5.6\% points (0.7601 vs 0.720).
	\item On \textbf{ISIC 2018}, the gap is even larger. Our 0.9261 accuracy is 3.2\% higher than FG-SSL's 0.897, and our 0.8706 F1-score is 6.7\% higher than their 0.816.
\end{itemize}
This improvement validates our hypothesis that chaotic map augmentation combined with attention fusion is a superior strategy for these fine-grained medical classification tasks.
\begin{table*}[!htpb]
	\centering
	\caption{Comparison with literature results (as reported in \cite{ref_jigsaw})}
	\label{tab:sota_comparison}
	\begin{tabular}{@{}llcc@{}}
		\toprule
		Dataset & Method & Accuracy & F1-Score \\
		\midrule
		\textbf{APTOS 2019} & CE \cite{ref_jigsaw} & 0.812 & 0.608 \\
		& CL \cite{marrakchi2021fighting} & 0.825 & 0.652 \\
		& ProCo \cite{yang2022proco} & 0.837 & 0.674 \\
		& FG-SSL \cite{ref_jigsaw} & 0.858 & 0.720 \\
		& \textbf{Chaos-SSL (Ours)} & \textbf{0.8726} & \textbf{0.7601} \\
		\midrule
		\textbf{ISIC 2018} & CE \cite{ref_jigsaw} & 0.850 & 0.716 \\
		& CL \cite{marrakchi2021fighting} & 0.865 & 0.739 \\
		& ProCo \cite{yang2022proco} & 0.887 & 0.763 \\
		& FG-SSL \cite{ref_jigsaw} & 0.897 & 0.816 \\
		& \textbf{Chaos-SSL (Ours)} & \textbf{0.9261} & \textbf{0.8706} \\
		\bottomrule
	\end{tabular}
\end{table*}
\begin{figure}[!htpb]
	\centering
	\includegraphics[width=0.9\columnwidth]{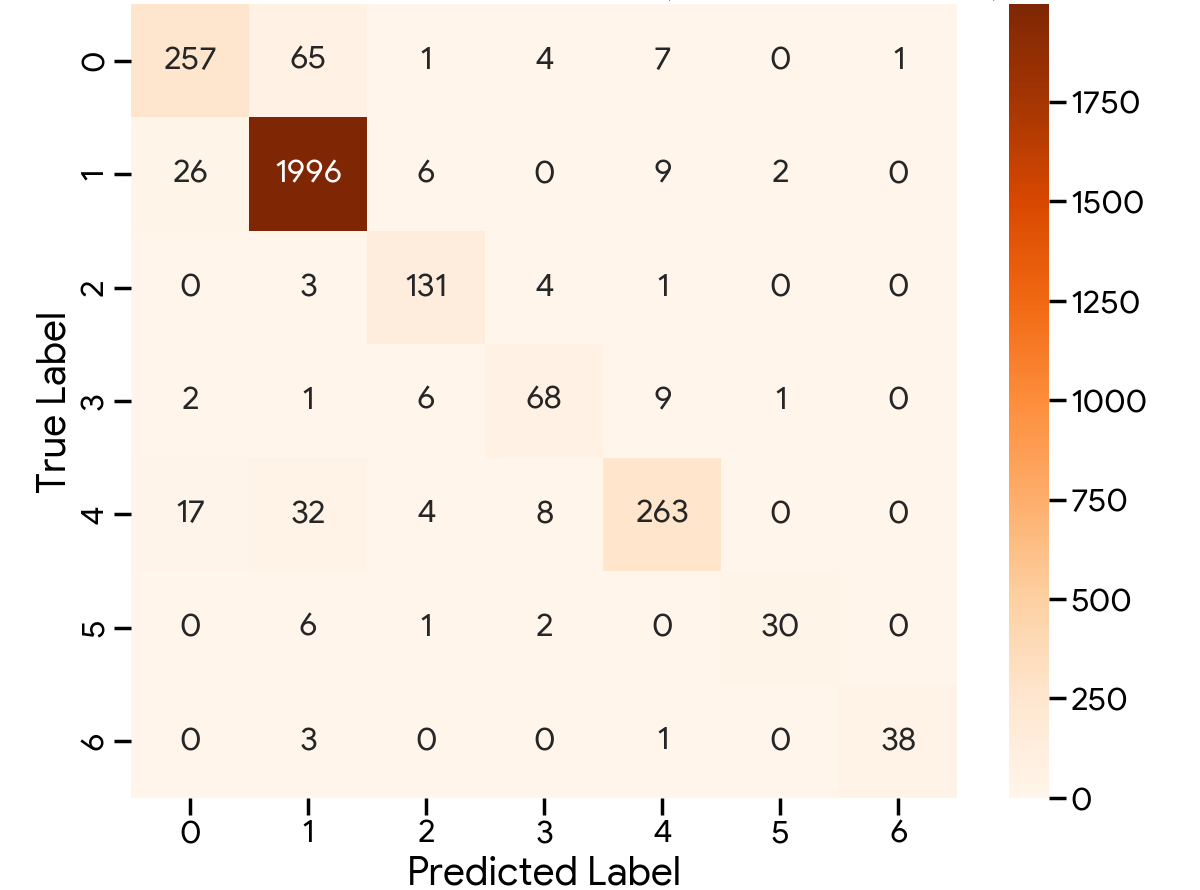}
	\caption{Confusion matrix for the best-performing model on the ISIC 2018 dataset (F1-Score: 0.8706). This corresponds to the 30-epoch Tent map SSL pre-training.}
	\label{fig:isic_cm}
\end{figure}

\section{\uppercase{Discussion}}
The results presented in Section IV suggest that our proposed Chaos-SSL framework is effective. We now discuss \textit{why} it works, its limitations, and directions for future research.

\subsection{Why Does Chaotic Augmentation Work?}
Standard augmentations (e.g., flip, jitter) teach invariance to changes in an image that do not alter its semantic content. Our ChaoticTransform is fundamentally different. It is a \textit{non-linear, deterministic, and complex textural distortion}. By creating a contrastive pair between a standard view ($v_1$) and a chaotic view ($v_{chaos}$), we are posing a much ``harder'' problem for the network: ``Find the core, invariant features of this pathology, even when its fine-grained texture is warped by a complex chaotic function.''

This forces the ConvNeXt-Tiny (Model 2) to become a specialist in ``textural and topological invariance.'' It must learn to ignore spurious, high-frequency noise introduced by the map and focus on the underlying, persistent structure of the pathology. This is in contrast to most contrastive approaches, which focuses on learning ``spatial relationships'' in the pretext task. Our results suggest that for fine-grained medical images, learning textural invariance (Chaos-SSL) is a more powerful and effective pre-training signal than learning spatial arrangement.

The ``Tent map's'' superior performance is intriguing. As a piecewise-linear map, its distortions may be more structured and ``edge-like'' compared to the smoother Sine map or the more aggressive Logistic map, making it an ideal ``hard'' augmentation that doesn't completely destroy the underlying information.

\subsection{The Power of Attention-Based Fusion}
Our results are not just from the Chaos-SSL model alone; they are from the fusion ensemble.
The ConvNeXt-Large (Model 1) provides a powerful, robust understanding of global shape, form, and context, learned from 1.2 million natural images. The ConvNeXt-Tiny (Model 2) provides a highly specialized, fine-grained understanding of medical textures. The feature fusion model acts as a learned ``arbitrator.'' Its attention mechanism dynamically learns to weigh these two feature streams, likely focusing on Model 1's features for global context (e.g., ``is this a skin lesion or an artifact?'') and Model 2's features for fine-grained classification (e.g., ``is this melanoma or a benign nevus?'').

\subsection{Loss Ablation Study}
``Table 3'' in \cite{ref_jigsaw} provides a salient comparison of loss functions. The authors compare a KL-Divergence loss against the Barlow-Twins loss, with the latter forming their final model. In Table \ref{tab:loss_comparison}, we expand this comparison to include our final model. This shows that our combined approach of a chaotic contrastive loss (NT-Xent) followed by an attention-fusion fine-tuning yields a substantial improvement over both loss function strategies on the ISIC 2018 dataset.
\begin{table}[!htpb]
	\centering
	\caption{Loss Function Ablation, as reported in \cite{ref_jigsaw}}
	\label{tab:loss_comparison}
	\begin{tabular}{@{}lcc@{}}
		\toprule
		Method (on ISIC 2018) & Accuracy & F1-Score \\
		\midrule
		FG-SSL (w/ KL-Div.) & 0.843 & 0.782 \\
		FG-SSL (w/ Barlow-Twins) & 0.897 & 0.816 \\
		\textbf{Chaos-SSL (Ours)} & \textbf{0.9261} & \textbf{0.8706} \\
		\bottomrule
	\end{tabular}
\end{table}

\subsection{Limitations and Future Work}
Despite the suggestive results, our work has limitations:
\begin{enumerate}
	\item \textbf{Dataset Scope:} We tested on two 2D, single-modality datasets. The method's efficacy on 3D data (CT, MRI) or other modalities (e.g., histology) is a future research direction.
	\item \textbf{Computational Cost:} The final ensemble model requires running two backbones (Large and Tiny), increasing inference time. For clinical deployment, model distillation into a single, compact network would be necessary.
	\item \textbf{Interpretability:} While we have a hypothesis for \textit{why} the Tent map works best, it remains an empirical finding. A deeper theoretical analysis of the interaction between chaotic map properties and learned feature invariances is needed.
\end{enumerate}

Future work will proceed in three main directions:
\begin{itemize}
	\item ``Expanding Augmentations'': Exploring compositions of chaotic maps (e.g., $f(x) = f_{logistic}(f_{tent}(x))$), as suggested by recent work \cite{ref_8}, or applying maps in the frequency domain.
	\item ``Expanding Frameworks'': Applying ChaoticTransform to other SSL frameworks, such as Masked Autoencoders (MAE) or self-distillation (DINO).
	\item ``Expanding Domains'': Validating this approach on 3D medical tasks and in other data-scarce, fine-grained domains (e.g., satellite imagery, materials science).
\end{itemize}

\section{\uppercase{Conclusion}}
This paper addressed the critical challenge of learning fine-grained representations in data-scarce medical imaging. We introduced \textbf{Chaos-SSL}, a novel self-supervised framework that moves beyond simple geometric augmentations. By using chaotic maps as a complex, non-linear transformation for contrastive learning, we successfully pre-trained an encoder to be invariant to high-frequency textural distortions.

We further proposed an attention-based fusion model that effectively combines the specialized, fine-grained features from our Chaos-SSL model with the robust, general-purpose features of a large ImageNet model.

Our comprehensive experiments on the ISIC 2018 and APTOS 2019 datasets showed that our method, using a 30-epoch Tent map pre-training, achieves results competitive with the state-of-the-art. It significantly outperforms prior SSL methods, including the recent FG-SSL approach, on both accuracy and F1-score. This work validates the use of chaotic dynamics as a powerful new tool in the self-supervised learning-based data augmentation, opening new avenues for developing highly specialized models for complex medical image analysis.

\section*{\uppercase{Acknowledgements}}

During the preparation of this work the authors used Google Gemini in order to improve readability and language. After using this tool, the authors reviewed and edited the content as needed and take full responsibility for the content of the publication.


\end{document}